\documentclass[conference]{IEEEtran}
\IEEEoverridecommandlockouts
\usepackage{cite}
\usepackage{amsmath,amssymb,amsfonts}
\usepackage{algorithmic}
\usepackage{graphicx}
\usepackage{textcomp}
\usepackage{xcolor}
\usepackage{booktabs}
\usepackage[ruled,vlined]{algorithm2e}
\usepackage{soul}
\def\BibTeX{{\rm B\kern-.05em{\sc i\kern-.025em b}\kern-.08em
    T\kern-.1667em\lower.7ex\hbox{E}\kern-.125emX}}

\newcommand\ourmethod{\textsf{AdaptAV}} 

\usepackage[letterpaper, top=0.7in, bottom=0.9in, left=0.6in, right=0.6in]{geometry}

\begin{document}

\title{AdaptAV: Continuous Adaption of Vision Models for Autonomous Vehicles Using Cloud-based Oracle}

\author{\IEEEauthorblockN{Yuheng Zhu, Dhruva Ungrupulithaya, Boluo Ge, Man-Ki Yoon}
\IEEEauthorblockA{\textit{Department of Computer Science}\\
\textit{{North Carolina State University}}}}

\renewcommand{\baselinestretch}{0.96}

\maketitle
\pagestyle{plain}

\begin{abstract}
Deploying vision perception models in autonomous vehicles requires that we prioritize inference speeds, resulting in a model with shallower architectures and lesser model parameters (i.e., more pruned). Such small models do not generalize well, which could result in poor performance when encountered with novel scenarios. We propose a system that overcomes this by continuously retraining the vision models on the cloud with data uploaded by vehicles. We leverage the abundant compute resources, including machine learning accelerators, of the cloud to run a state-of-the-art, highly accurate \emph{oracle} model that will guide the retraining process of the on-vehicle model. This newly trained model is transmitted to the vehicle over the network and is utilized by the vehicle for perceptions, leading to improved inference accuracy over time.
\end{abstract}

\section{Introduction}

The proliferation of autonomous vehicles (AVs) has been one of the most significant trends in vehicular technology with 
automakers investing heavily \cite{lipson2017driverless}. 
Perception models are key to AVs' success and work by taking inputs from an array of sensors like cameras and LiDARs to assess the surrounding environment and make relevant decisions. The development of these intelligent algorithms is highly data-driven and is a result of the collection of millions of hours of driving data in a variety of unique environmental conditions and driving scenarios and training the models on these extensive datasets. 

\begin{figure}[t]
    \centering
    \includegraphics[width=0.85\columnwidth]{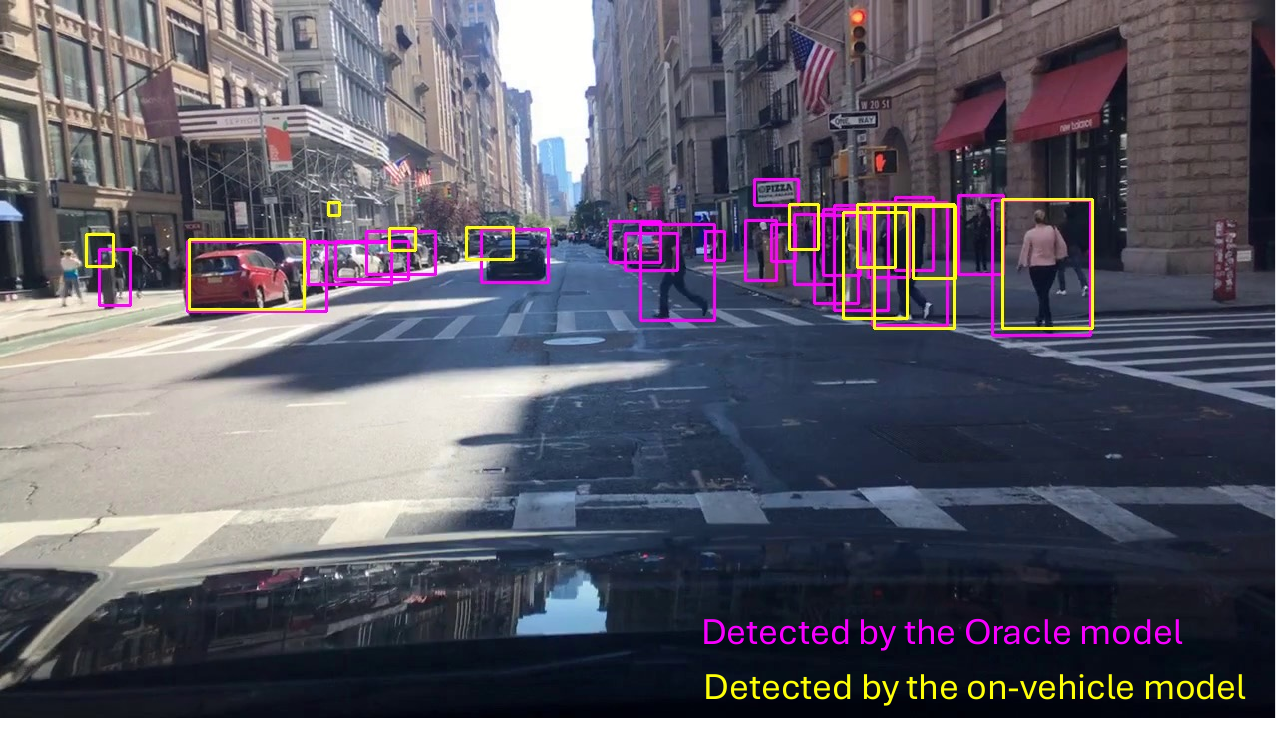}
    \vspace{-0.3cm}
    \caption{A scenario where the vision model deployed in the vehicle fails to detect multiple important objects. The purple bounding boxes correspond to the detection of the larger Faster R-CNN model, and the yellow bounding boxes correspond to the smaller SSD-MobileNet model. 
    }\vspace{-0.3cm} 
    \label{fig:accident}
  \end{figure}

Although the models are tested vigorously to ensure that they perform well in a variety of challenging conditions, optimal performance is not always guaranteed. A poor performance can be ascribed to two primary reasons: a sub-optimal model and/or a lack of comprehensive training. Due to the constrained computational and memory resources available in vehicles, the perception models running in these systems may not represent the 
most precise models currently accessible. These models are instead optimized for resource efficiency, resulting in the model performing inference quickly to meet the real-time requirements of an autonomous vehicle. Thus, on encountering a novel and challenging scenario, which might not be a part of the training dataset, the perception model performs poorly, leading to inferior decisions. For instance, Fig. \ref{fig:accident} shows the detection of a larger model (Faster R-CNN~\cite{ren2015faster}) and that of a smaller model (SSD-MobileNet~\cite{howard2017mobilenets}\cite{liu2016ssd}) that would run on the vehicle, both of which are trained on the Waymo Open Dataset~\cite{Sun_2020_CVPR_Waymo}. There is a significant difference in the inference outputs of the two models, with the smaller MobileNet model missing some key objects such as the pedestrian crossing the road. This poor perception performance, if not addressed, may lead to catastrophic consequences\cite{UberCrash}\cite{valdes2024waymo}. 

The conventional approach would be to deploy test vehicles to collect more data and use it to train and develop a better perception model. This improved model would then be updated when the vehicle comes to the service center as a software update, or if supported, through an over-the-air (OTA) update. However, the process of data collection and development of the updated model would take significant amounts of time and resources and still may not address all the shortcomings of the previous model~\cite{teslaOTA}.

\begin{figure*}[t]
    \centering
    \includegraphics[width=0.7\textwidth]{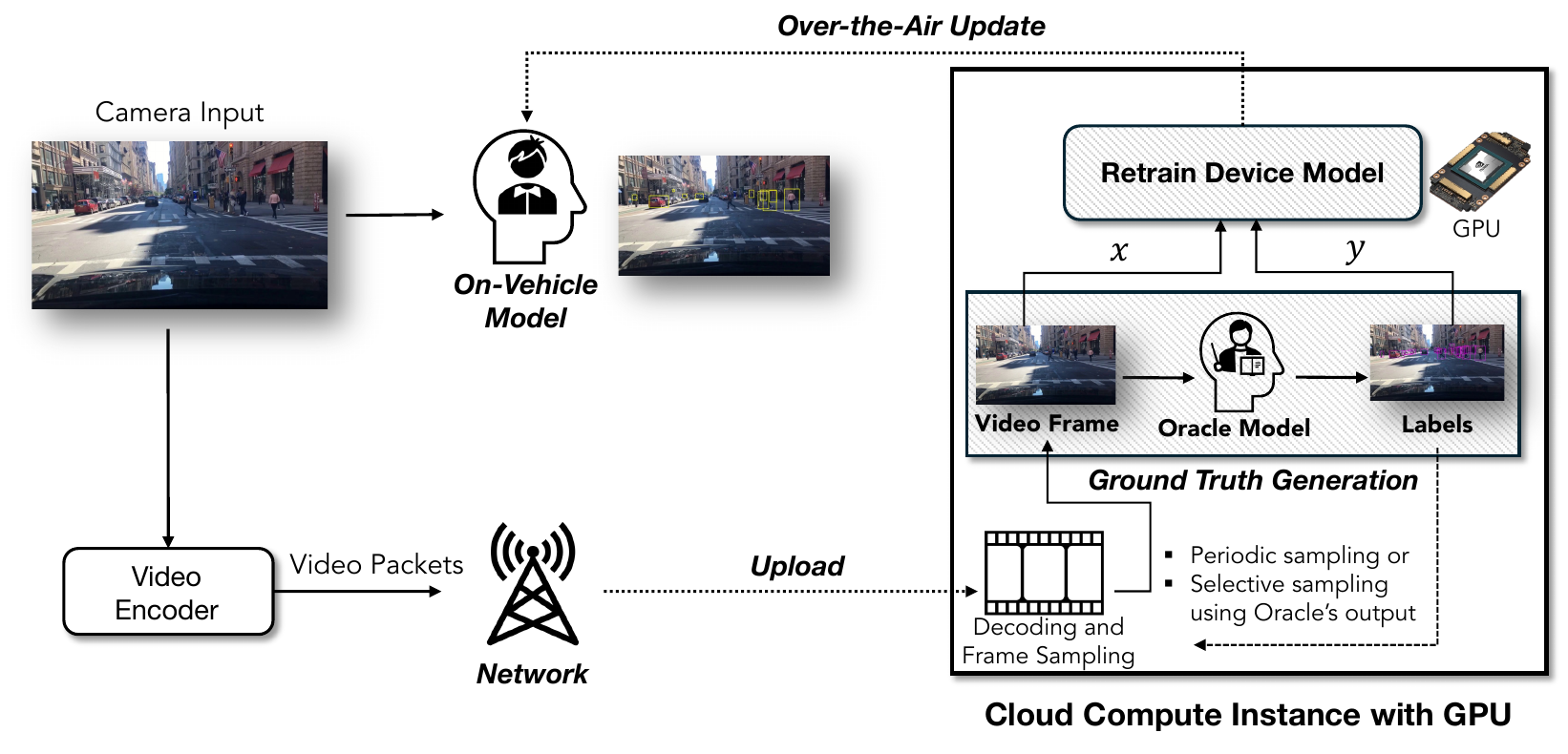}
    \vspace{-0.3cm}
    \caption{\ourmethod{} continuously updates the vision model running on the vehicle. The image stream used by the vision model is compressed and uploaded to the cloud. A cloud instance with powerful compute with machine learning accelerators uses the data to retrain the model. The retrained model is then deployed back to the vehicle, closing the loop.} 
    \vspace{-0.5cm}
    \label{fig:sys_diag}
\end{figure*}

Drawing from the above insights, we propose \ourmethod{}, a closed-loop software architecture that continuously improves and updates the model running in the vehicle. We achieve this by leveraging the power of computing on the \emph{cloud}, which enables us to apply a complex, powerful perception model to flag erroneous or poor decisions of the on-vehicle model. These \emph{oracle} models are developed to prioritize inference accuracy rather than speed, which allows us to use them to steer the (re)training of the small model that runs on the vehicle. The vehicle continuously logs the vision data and pushes it to the cloud in a bandwidth-efficient manner by using popular video compression techniques, allowing us to deploy this system even with limited network connectivity.

In this paper, we make the following contributions:

\begin{itemize}
    \item We present \ourmethod{}, a closed-loop software framework that continuously updates the vision model running in the vehicle using the cloud-side oracle model and the most recently captured image stream;
    \item We propose frame sampling techniques that can reduce overfitting of the vision model;  
    \item We analyze the \ourmethod{} design parameters and their impacts on ML inference quality;
    \item We show the practicality of \ourmethod{} by applying it to a prototype vehicle integrated with a public cloud.
\end{itemize}

\section{Related Work}

For visual object detection, existing methods can be divided into one-stage approach, represented by YOLO\cite{redmon2018yolov3}\cite{bochkovskiy2020yolov4}\cite{li2022yolov6}, SSD\cite{liu2016ssd}, and two-stage approaches represented by R-CNN and its various variants\cite{girshick2015fast}\cite{ren2015faster}\cite{he2017mask}. One-stage approaches can generate object location and confidence in only a single stage, which ensures the inference speed and thus is more suitable to be deployed on vehicles. Two-stage models need to generate region proposals first and then generate final detection based on the region proposals. It guarantees relatively better accuracy than the one-stage models but at the cost of higher latency.

One common approach to use cloud computing resources to alleviate computational power constraints of device is to apply early exit strategy on the device model, which reduces the device-side inference latency by implementing some scheduling procedures to split some layers of the inference model to the cloud or edge computing platform\cite{jeong2018ionn}\cite{laskaridis2020spinn}. For example, Edge AI\cite{li2019edge} implements a DNN-based collaborative inference framework for edge computing, which uses early exit to distribute computing loads across devices and edge servers, but such approaches rely on consistent model architectures, making the framework overly coupled with the inference model architecture. %This will pose a challenge for future technology updates.

The other approach is that the device model completely undertakes all visual inference tasks for the vehicle and uses the cloud model as a teacher to transfer its generalization capability to the device model. This kind of \emph{knowledge distillation} approach ensures the decoupling of the model architectures between the device and the cloud. However, it requires the visual data collected on the vehicles to be sent to the cloud for inference, which leads to higher demand on network bandwidth. 
EdgeCompression \cite{lu2020edge} proposes the use of compressive imaging (CI) cameras in addition to specialized energy-efficient deep neural networks in the cloud and edge servers. However, this approach would require the usage of less prevalent and expensive CI cameras and would also require a complete redesign of the other vision-based software modules in the vehicle.

Ekya \cite{bhardwaj2022ekya} is a similar framework for continuous retraining of a vision model. But, it uses edge servers instead of the cloud because the bandwidth required to upload raw camera data is too high. More importantly, their edge servers also perform vision inference, unlike our approach, which performs inference locally on AV devices.
Khani et al. \cite{khani2021real} propose an on-the-device frame sampling technique to reduce the network bandwidth requirements. For the sampling, it utilizes the device model's results, which can be problematic since these results may not be accurate enough to help determine the samples relevant for the retraining process. 
Our framework is novel in that it performs inferences locally but leverages an oracle model running on a cloud platform to continuously improve vision models in autonomous vehicles.  

\section{AdaptAV: Continuous Adaption Using Cloud-based Oracle Model}
\subsection{High-level Idea of \ourmethod{}}
Vehicles are resource-constrained in terms of computing power, making the deployment of extremely large models with complex architectures infeasible. Although using these models for real-time inference in the vehicle is impractical, its high accuracy allows us to use it as an \emph{oracle}. The oracle model trained on an extensive dataset generalizes better than the smaller shallower models running on the vehicles, making them perform significantly better when dealing with novel scenes not seen during training. 
As seen in Table \ref{tab:oracle_vs_device}, the Faster-RCNN model (the \emph{oracle} model) has a significantly higher mean average precision (mAP)\footnote{It is an evaluation metric for object detection models that summarizes the precision-recall curve as the area under the curve across all confidence thresholds. The \emph{precision} of a model is calculated as TP/(TP + FP) where TP is the number of true positives, and FP is the number of false positives. The \emph{recall} of a model is calculated as TP/(FP + FN) where FN is the number of false negatives. A higher AP indicates a higher accuracy and vice versa.} of 52\% compared to the smaller SSD-Mobilenet model that runs in the vehicle (mean average precision of 23\%). Both models were trained on 35,712 training images of the Waymo Open Dataset \cite{Sun_2020_CVPR_Waymo} and were tested on a separate subset of 3,968 images.

\begin{table}[tb]
    \centering\footnotesize
    \caption{Mean average precision for oracle model and device model}\label{tab:oracle_vs_device}
\begin{tabular}{crr}
\toprule
    \textbf{Class}  & \textbf{Faster R-CNN} & \textbf{SSD-Mobilenet} \\
                 \midrule
\textbf{person}  & 0.45               & 0.17                      \\
\textbf{car}     & 0.72               & 0.49                      \\
\textbf{bicycle} & 0.41               & 0.18                      \\
\textbf{sign}    & 0.48               & 0.09                      \\
\midrule
\textbf{Average} & 0.52               & 0.23                     \\
    \bottomrule
\end{tabular}
\vspace{-0.4cm}
\end{table}

Drawing from these results, we hypothesize that the predictions from the oracle model can be treated as the \emph{ground truth} that we can use to retrain the on-vehicle device model.  \ourmethod{} is a closed-loop pipeline that streamlines this process as shown in Fig.~\ref{fig:sys_diag}. Autonomous vehicles typically have limited compute and memory resources that are dedicated to performing time and safety critical tasks essential to its functioning. Leveraging the power of cloud storage and computing gives us access to a near-infinite memory resource and extremely powerful computing capabilities, to easily implement \ourmethod{}. It continuously captures and streams the camera images to the cloud. The cloud-side compute instance runs the oracle model, which will consume the streamed camera data. The results of the oracle model are used as the ground truth to retrain the device model. The retrained model is then sent to the vehicle. \ourmethod{} hence continuously improves the on-vehicle model, making it more robust to novel scenarios.

\subsection{Video Compression}

As shown in Fig.~\ref{fig:sys_diag}, raw image stream is fed periodically to \ourmethod{} as they are captured by the camera.
Although the use of cloud resources offers significant advantages, 
the network bandwidth could easily be the bottleneck because 
transmitting the raw image stream to the cloud would be prohibitively slow. 
Camera sensors used in autonomous vehicles (such as MIPI CSI cameras) generate high-resolution images at a high rate to capture the finest details of the environment. Unlike webcams, they do not have built-in codecs, so the raw frames generate an enormous amount of data. 
Suppose the vehicle has a single 3-channel Full HD camera (1920 x 1080 x 3) recording at 30 Hz. Then, we would need a network capable of upload speeds of approximately 1.4 Gbps. 

\ourmethod{} tackles the challenges by compressing the raw frames into a video stream, which consumes much less network resources to transmit. Specifically, the raw frames are encoded into a sequence of video \texttt{packets}, which are queued and pushed to the cloud by a separate thread. For the compression, we use \texttt{FFmpeg} \cite{ffmpeg} as a backend, which allows us to use a variety of video codecs. In this paper, we employ the ubiquitous H.264 codec (\texttt{libx264}) for video compression. It is the most common encoding format used by many domains such as online streaming and digital television. Its compression ratios are also highly effective; for instance, the Waymo dataset that we use for our evaluation, consisting of 200 subsets, requires approximately 286 GB of storage. However, it can be compressed into a video of merely 4.2 GB in size, saving more than 98\%.

\begin{figure}[t]
    \centering
    \includegraphics[width=0.75\columnwidth]{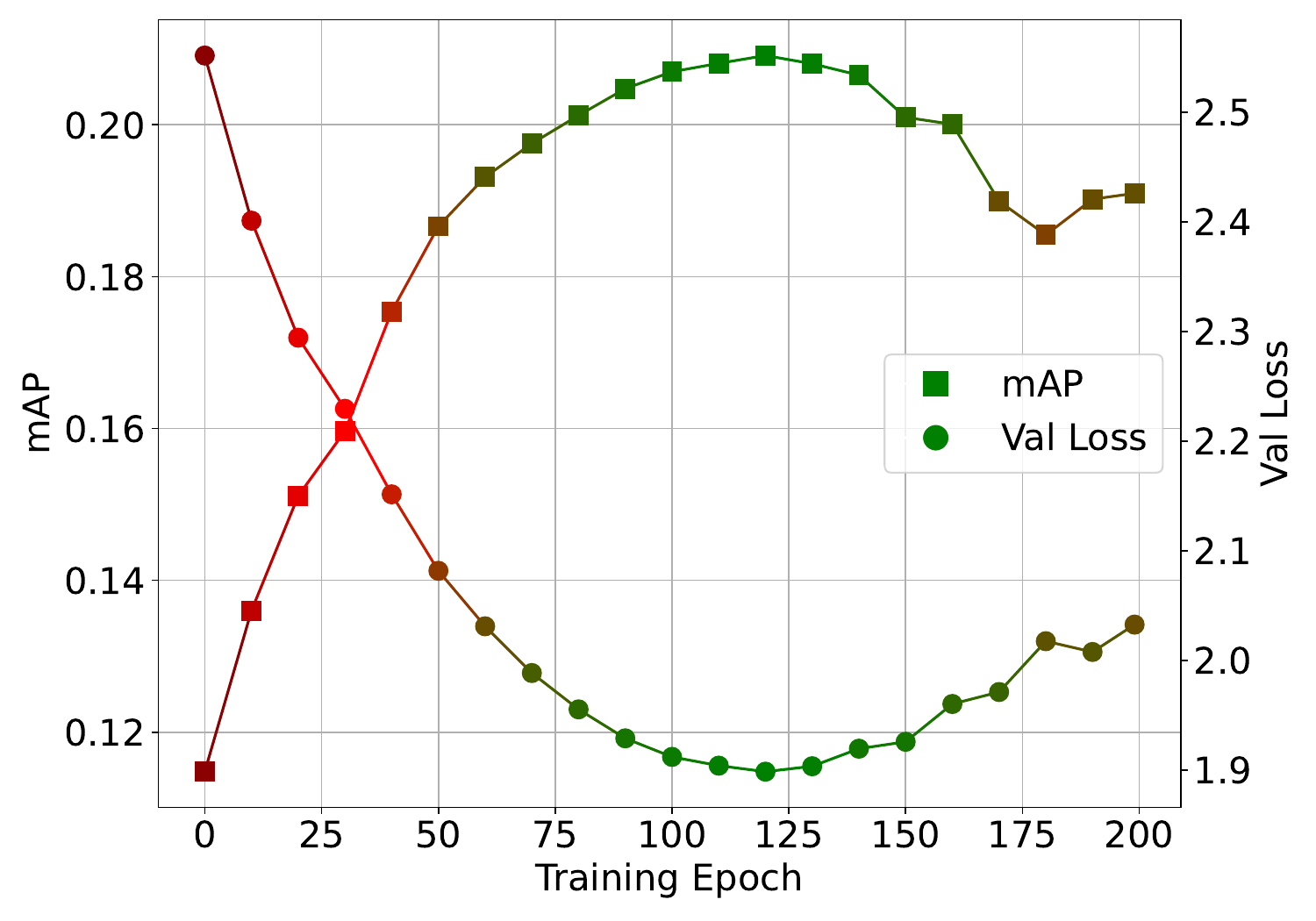}
    \vspace{-0.3cm}
    \caption{A model is likely to overfit when using frames captured/sampled at a high rate. Hence, a frame sampling is necessary.}
    \label{fig:overfitting_combined}
    \vspace{-0.5cm}
  \end{figure}

\subsection{Frame Sampling}

\begin{figure*}[t]
    \centering
    \includegraphics[width=0.68\textwidth]{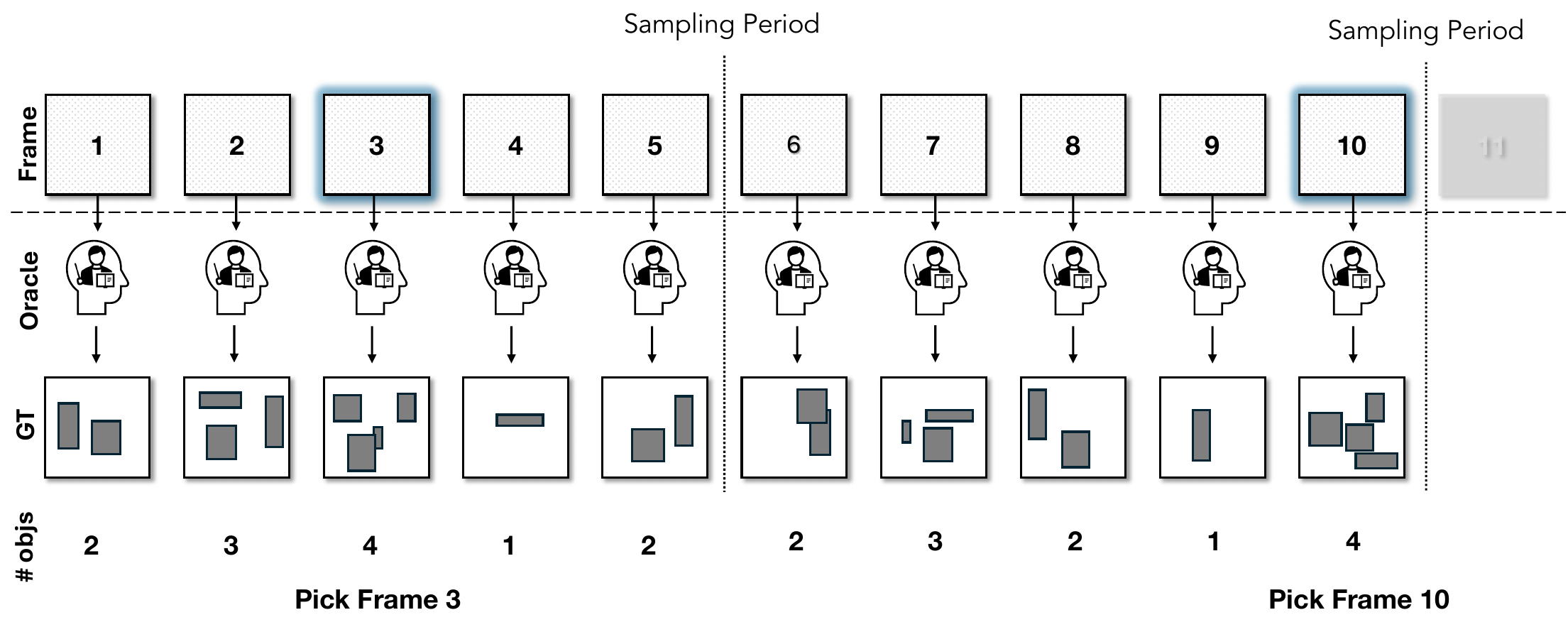}
    \vspace{-0.3cm}
    \caption{Density-based frame sampling. During each sampling period, the one that has the largest number of detected objects is selected.} 
    \label{fig:sampling}
    \vspace{-0.5cm}
\end{figure*}

\begin{figure*}[t]
    \centering
    \includegraphics[width=0.68\textwidth]{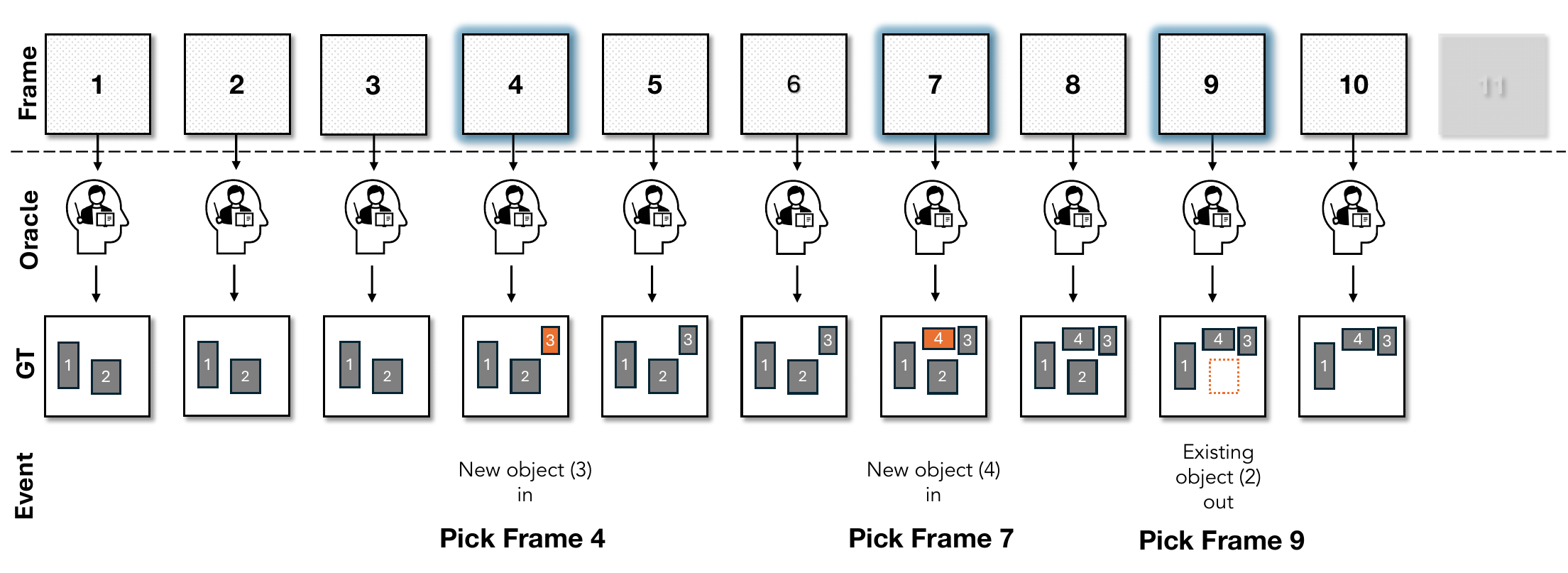}
    \vspace{-0.3cm}
    \caption{Event-based frame sampling. An event is defined by the appearance of a new object or the disappearance of an existing object.} 
    \label{fig:incident_sampling}
    \vspace{-0.5cm}
\end{figure*}

The compressed image stream transmitted to the cloud is extensive. If the camera is emitting data at a rate of 30 frames per second, we would have 108,000 images for just 1 hour of driving. Using all these images to re-train the device model would take a large amount of compute resources. Moreover, we risk \emph{overfitting} the model by using all the images, i.e., the model will \emph{memorize} rather than \emph{generalize}. Fig.~\ref{fig:overfitting_combined} shows that when using all of the BDD100K dataset (which will be introduced in the next section) for training, an overfitting starts to occur. To overcome the problem, we propose the following \emph{frame sampling} techniques.

\vspace{0.25\baselineskip}
\paragraph{Periodic sampling} This is the baseline approach in which we periodically sample an image in a sampling period across a chosen window of the recorded data. For instance, we can choose the first image for every $n$ images, effectively reducing the training dataset to $1/n$ of its original size compared to using all incoming images.

\vspace{0.25\baselineskip}
\paragraph{Density-based sampling} Another approach is to take into account the oracle model's prediction from each of the images. As an example, particularly for object detection tasks, we propose a method that selects the frame with the \emph{highest} number of objects in every sampling period, as illustrated in Fig.~\ref{fig:sampling}. If all the frames within a period have the same number of objects, one is randomly chosen. The number of objects in each image is obtained from the prediction by the oracle model. The idea behind this approach is that for a sequence of images captured within a short period of time, their objects and predictions should be similar. If we want to downsample this set of images, the most rational way to do so is by selecting the frame with the \emph{most} information, which can be implied by the number of objects present. 

\vspace{0.25\baselineskip}
\paragraph{Event-based sampling} 
% In autonomous driving, it is crucial to maintain the model's performance during unexpected situations and critical events. 
For event-based sampling, the criterion for selecting a frame is the occurrence of an \emph{event}, which can be defined based on changes in the types or behavior of detected objects. For example, as Fig.~\ref{fig:incident_sampling} shows, the occurrence of a new object not previously present or the disappearance of an existing object is considered an event. Similar to the density-based sampling, this approach also utilizes the oracle model's predictions, incorporating not only object detection results but also \emph{tracking} information. Frames are selected when the model detects preset events, including when a new object appears and an existing object disappears. The rationale behind the idea is that when the same objects appear in a series of frames without any significant change, they do not deliver much `new' information. This sampling approach is designed to ensure that only frames with \emph{meaningful changes} are selected, maintaining the relevance of the sampled frames. 
In our implementation, we use \textsf{SORT}\cite{Bewley2016sort}, a real-time tracking algorithm that tracks each unique object in continuous frames and generates a unique identifier for each of them. By using it to track the motion of objects in frames, we detect when each object first enters and leaves the frame and mark it as an event.

If available, one could instead detect events based on other sensor inputs, such as vehicle turns, sharp braking, or the presence of other objects at relatively high speeds. The proposed sampling approaches are critically different from existing techniques such as \cite{khani2021real} in that we leverage the \emph{oracle} model's results, not the device model. Using the device model's inference outputs risks relevant objects going undetected or being misclassified during the retraining process.

\vspace{0.25\baselineskip}
Note that \emph{filtering} differs from sampling. For instance, Reducto \cite{li2020reducto} is an image filtering mechanism running directly on a camera system-on-chip that filters out irrelevant frames for the perception models at the camera level. Mistakes made by the filtering algorithm may drop camera frames for a long period of time, which can be a critical hazard to safety-critical systems like autonomous vehicles.

\section{Evaluation}

\subsection{Setup}

\subsubsection{Platforms}

Our implementation is based on an NVIDIA AGX Orin Development Kit \cite{JetsonOrin} that we run on a small-footprint vehicle (approx. 3 ft x 6 ft) shown in Fig.~\ref{fig:ecoprt_sys_diag}. It 
features a 
12-core ARM Cortex-A78AE CPU, each core running at 2.2 GHz, a 2048-core NVIDIA Ampere GPU, and 64 GB of memory. The software system is configured with NVIDIA JetPack SDK v5.1.2, which includes Linux for Tegra 35.4.1,
CUDA v11.4, cuDNN v8.6.0, and TensorRT v5.1.2. We use a workstation with an Intel i9-13900F CPU, 64 GB memory, and NVIDIA GeForce RTX 4090 GPU for in-lab evaluation with open-source datasets. It operates Ubuntu v22.04.4 LTS system with kernel v6.5.0, and CUDA v12.3. For the prototype unmanned vehicle (detailed in Section~\ref{sec:usecase}), we utilize Google Cloud \cite{gcloud} Storage Bucket 
where video packets are uploaded to. We use a Google Cloud virtual machine instance \texttt{a2-highgpu-1g} for running the Faster R-CNN oracle model and retraining the on-device model. 
It
has 12 virtual CPUs, 85 GB of memory, an NVIDIA A100 (40 GB) GPU, and operates Debian 11.1 and CUDA v12.2.

\vspace{0.25\baselineskip}
\subsubsection{Datasets}

\begin{itemize}
    \item Waymo Open Dataset\cite{Sun_2020_CVPR_Waymo}: We choose the first 200 driving video records out of the 2D object detection validation set from Waymo Open Dataset v1.4.2. The subset consists of 39,681 frames with a resolution of 1920x1280 captured at 10 Hz. %It is used for the parameter initialization of our detection models.
    It is used only for the \emph{initial} training of our detection models.

    \vspace{0.25\baselineskip}
    \item BDD100K\cite{bdd100k}: MOT-20202 Images set from BDD100K serves as a simulation of actual video stream. The frames  
    are captured at 5 Hz with 1280x720 resolution. The dataset comprises 200 driving video records (approx. 200 frames per record), amounting to a total of 39,973 frames.
\end{itemize}

\vspace{0.25\baselineskip}
\subsubsection{Models}

\begin{itemize}
    \item Faster R-CNN\cite{ren2015faster}: It
    is an object detection model that uses a region proposal network to classify object proposals and predict bounding boxes within a single pass through the network. We utilize it as the \emph{oracle} model 
    for generating the most accurate detection results possible. The model contains 41,092,136 parameters, and the model file occupies 161,916 KB of disk space.
    \vspace{0.25\baselineskip}
    \item SSD-MobileNet\cite{howard2017mobilenets}\cite{liu2016ssd}: 
    It 
    is a fast and efficient object detection model that combines Single Shot Multibox Detector with a MobileNet architecture, making it ideal for real-time applications on resource-constrained devices. In our work, it is deployed on the vehicle to perform real-time detection on camera inputs as the \emph{device} model as mentioned earlier. The model contains 3,206,976 parameters, and the model file size is 27,641 KB.
\end{itemize}

\vspace{0.25\baselineskip}
On the Orin device, while the SSD-MobileNet takes approximately 5 ms to make a single inference, the Faster R-CNN model takes 130 ms. This shows that the oracle model is too large to be run on the on-vehicle device.

\subsection{Results}

We evaluate \ourmethod{} on the BDD100K object detection dataset mentioned above. The dataset is split into the training set (90\%) and the validation set (10\%). Recall that both the \emph{oracle} model on the cloud and the \emph{device} model deployed on the vehicle are pre-trained with the Waymo dataset and thus have never seen the BDD100K dataset.

For the retraining process, we consider the following two hyperparameters that affect the retraining speed and quality:
\begin{itemize}
    \item \textbf{Retraining sample rate}: Cameras with a high sampling rate will produce too many images, which can result not only in an elongated retraining process but also overfitting, as discussed earlier. Hence, it is necessary to downsample the acquired images to reduce the size of the retraining set.

    \item \textbf{Retraining window size}: It represents the interval between successive retrainings. Similar to the retraining sample rate, a retraining window size that is too large will lead to unnecessarily long retraining times, while a size that is too small may result in the model failing to accurately learn features within the corresponding retraining window.
\end{itemize}

\begin{figure}[t]
    \centering
    \includegraphics[width=0.8\columnwidth]{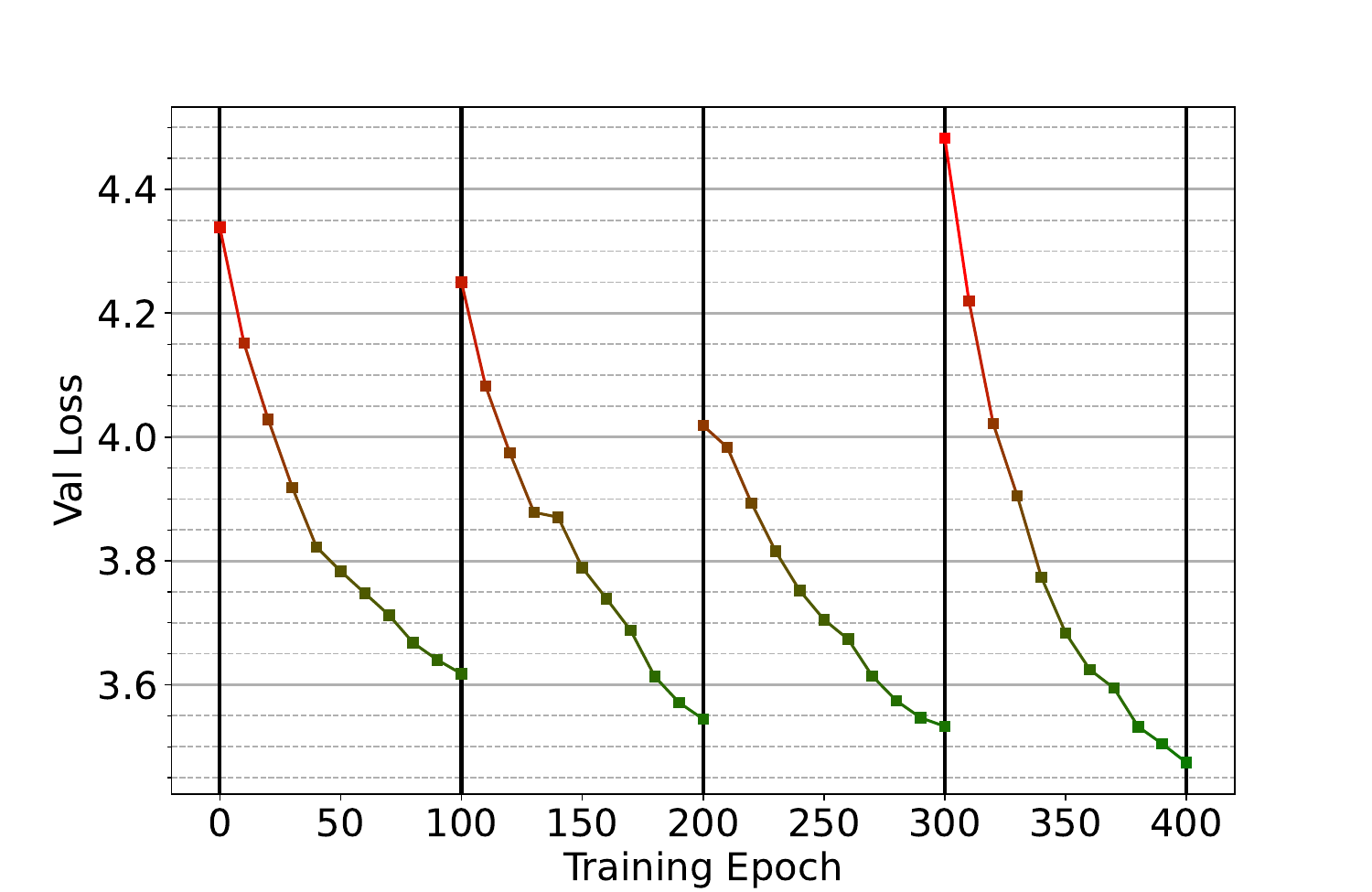}
    \vspace{-0.3cm}
    \caption{The validation loss during the continuous retraining of the SSD-Mobilenet model. There are a total of 4 retraining windows.  } 
    \label{fig:first_val_loss_cascade}
    \vspace{-0.5cm}
  \end{figure}

\begin{figure}[t]
    \centering
    \includegraphics[width=0.8\columnwidth]{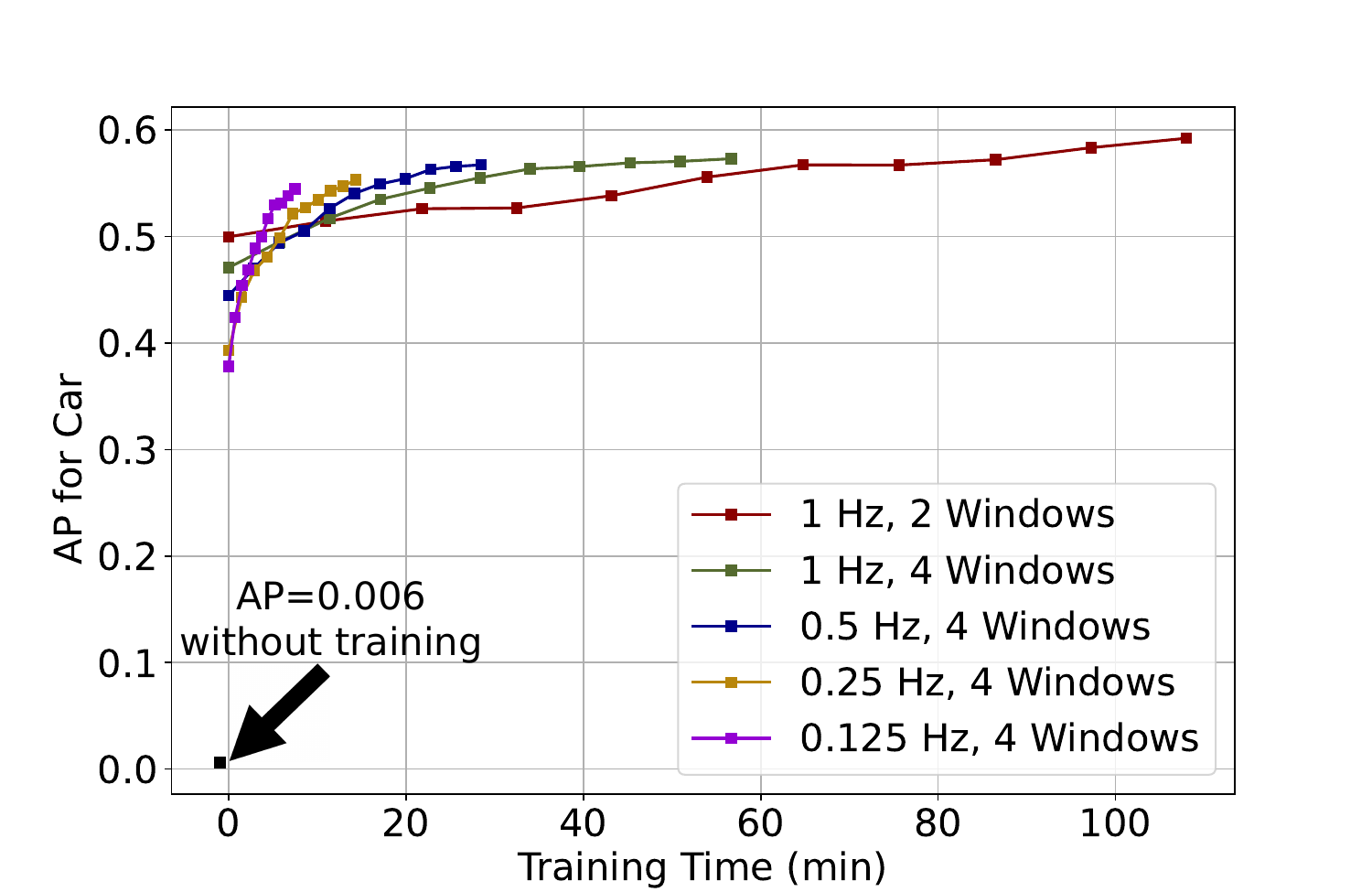}
    \vspace{-0.3cm}
    \caption{Model accuracy (AP for the \textsf{car} class) v.s. retraining time for different hyperparameter configurations. The number of windows refers to the number of retraining windows into which the BDD100K training set is divided equally. Each line is cut at the point where the y-axis value converges.}
    \label{fig:joint_car_mAP}
    \vspace{-0.5cm}
  \end{figure}

\vspace{0.25\baselineskip}
As mentioned above, we use the BDD100K dataset for the continuous retraining of the SSD-Mobilenet model. We first equally divide the BDD100K dataset into four retraining windows. Hence, each window consists of about 9000 images; one can imagine the on-vehicle camera capturing 9000 images. Then, \ourmethod{} applies a frame sampling technique, which defaults to periodic sampling, and then trains the model to convergence on each window using the preset retraining hyperparameters. 
Fig.~\ref{fig:first_val_loss_cascade} shows the validation losses over the whole retraining steps. Note that at the end of each retraining window, the updated model is deployed (updated to the device). Hence, what will be visible from the on-vehicle model are the final loss values at the end of each window (i.e., 100th, 200th, 300th, 400th). As we can see, the loss of the model decreases as the retraining process repeats, which indicates that its detection accuracy improves with the continuous retraining. This is further emphasized in Fig.~\ref{fig:joint_car_mAP}, where the accuracy of the model (y-axis) when directly tested on the BDD100K dataset (represented by a single black point) is a mere 0.6\%. The model, having never encountered the scenarios in the BDD100K dataset, performs extremely poorly. Retraining the model by just a single epoch, represented by the data points at a time near 0 min (x-axis), leads to a drastic improvement in the model performance, with continuous retraining ultimately yielding an AP of nearly 60\%.

Fig.~\ref{fig:joint_car_mAP} also evaluates the effect of different retraining window sizes on the model retraining. As can be seen, although the model can converge quickly even with a very low sampling rate (small training set) the accuracy decreases; on the contrary, a larger training set obtained with a higher sampling rate results in a higher accuracy of the model but the time required for convergence is also longer. In practice, one needs to select proper hyperparameters according to the computing resources on the cloud and the sampling rate of the vehicle camera, so that the frames for retraining are not sampled too fast or slow.

  \begin{figure}[t]
    \centering
    \includegraphics[width=0.8\columnwidth]{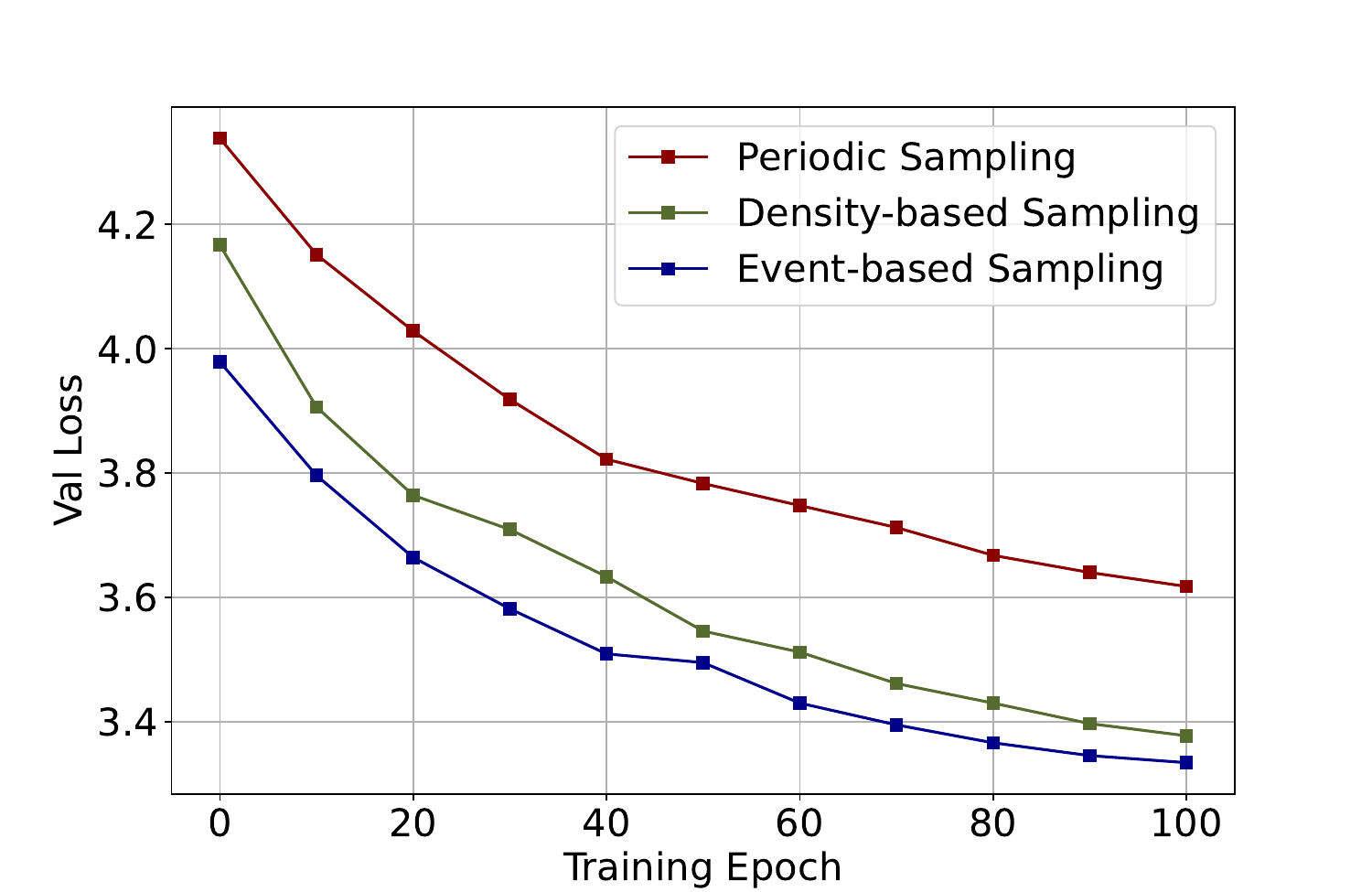}
    \vspace{-0.3cm}
    \caption{The impacts of different sampling approaches on the validation loss.\vspace{-0.35cm}} 
    \label{fig:samplings}
    \vspace{-0.3cm}
  \end{figure}

    \begin{figure}[t]
    \centering
    \includegraphics[width=0.9\columnwidth]{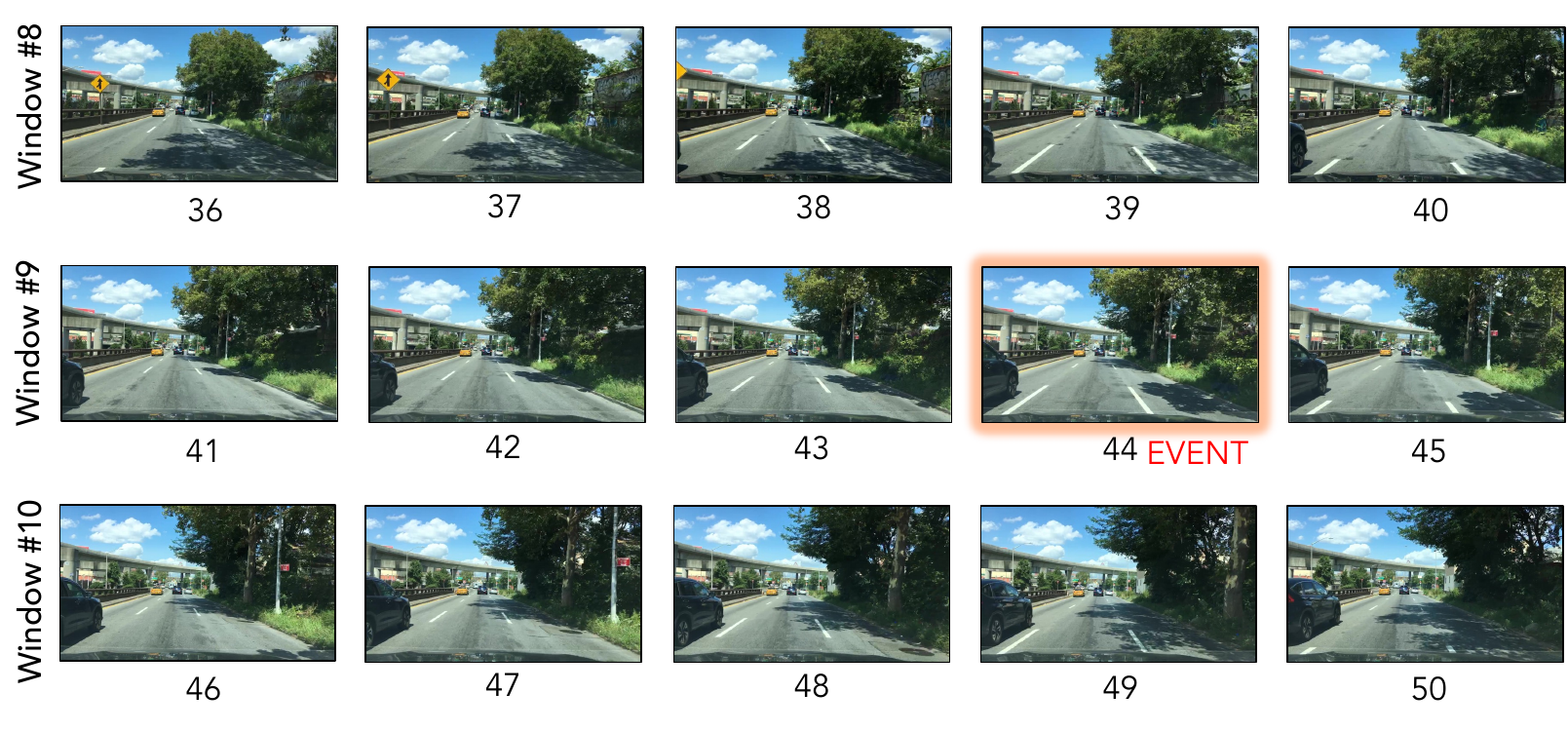}
    \vspace{-0.3cm}
    \caption{The blue-colored vehicle on the left of the scene is first detected by the Oracle model at Frame 44. Thus, it is the only frame sampled by the event-based method. The density-based method samples three frames from these three windows, each with a length of five, although the scenes are similar.} 
    \label{fig:sampling_frames}
    \vspace{-0.3cm}
  \end{figure}

Fig.~\ref{fig:samplings} shows the impacts of periodic sampling, density-based sampling, and event-based sampling on the validation loss during retraining. Compared to the periodic sampling, the two selective samplings help improve the model accuracy. As discussed in the previous section, these selective sampling techniques improve the quality of the frame samples by selecting the ones that contain more/better information than others.
Meanwhile, the event-based sampling performs the best out of the three approaches overall. 
We can attribute this result to the fact that both periodic and density-based methods are forced to sample \emph{at least} one frame during each sampling period. This can result in sampling \emph{redundant} frames, such as when the scene is static or has low dynamics (e.g., due to the ego vehicle moving slowly or being stopped), which leads to consecutive periods sampling very similar frames. Fig.~\ref{fig:sampling_frames} illustrates such a situation. 
These redundant frames contribute little new information during retraining. Conversely, in high-dynamic scenes, both periodic and density-based methods are limited to sample \emph{at most} one frame during each sampling period. As a result, they may miss frames that would have been informative if sampled. The event-based sampling is not constrained by the sampling period; it can sample more or fewer frames depending on the scene dynamics, which makes it suitable for mobile systems like vehicles. 

It is important to note that both the density and event based sampling techniques overcome the overfitting issue that is prevalent with continuous training of models. However, implementing the event-based sampling requires additional computing resources to run an object-tracking algorithm. Hence, one may want to opt for the density-based method if computing power is limited. Although it is not as powerful as the event-based method, it performs better than the naive periodic sampling technique while also preventing overfitting.

\subsection{Usecase}\label{sec:usecase}

\begin{figure}[t]
    \centering
    \includegraphics[width=0.9\columnwidth]{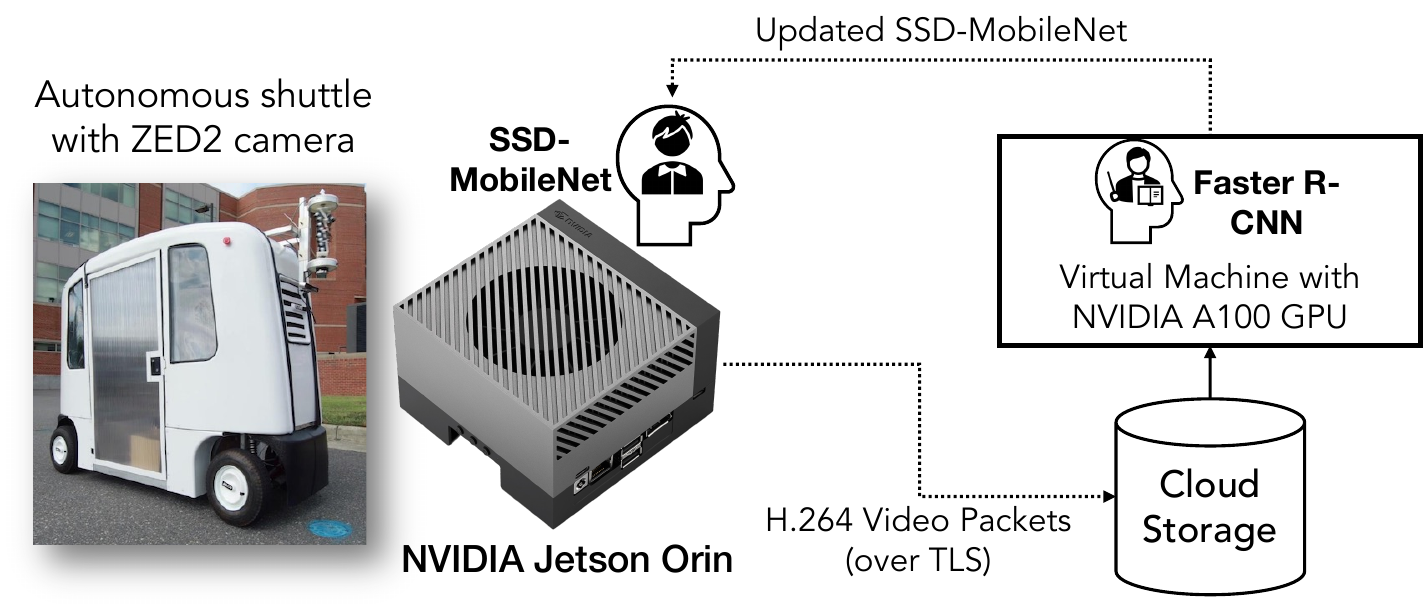}
    \vspace{-0.3cm}
    \caption{\ourmethod{} implemented on a small autonomous shuttle interacting with a public cloud platform. The vehicle uses the NVIDIA Jetson Orin as its computing platform and uses a ZED2 camera inputs for its vision tasks.} 
    \vspace{-0.3cm}
    \label{fig:ecoprt_sys_diag}
\end{figure}

\begin{figure}[t]
    \centering
    \includegraphics[width=0.8\columnwidth]{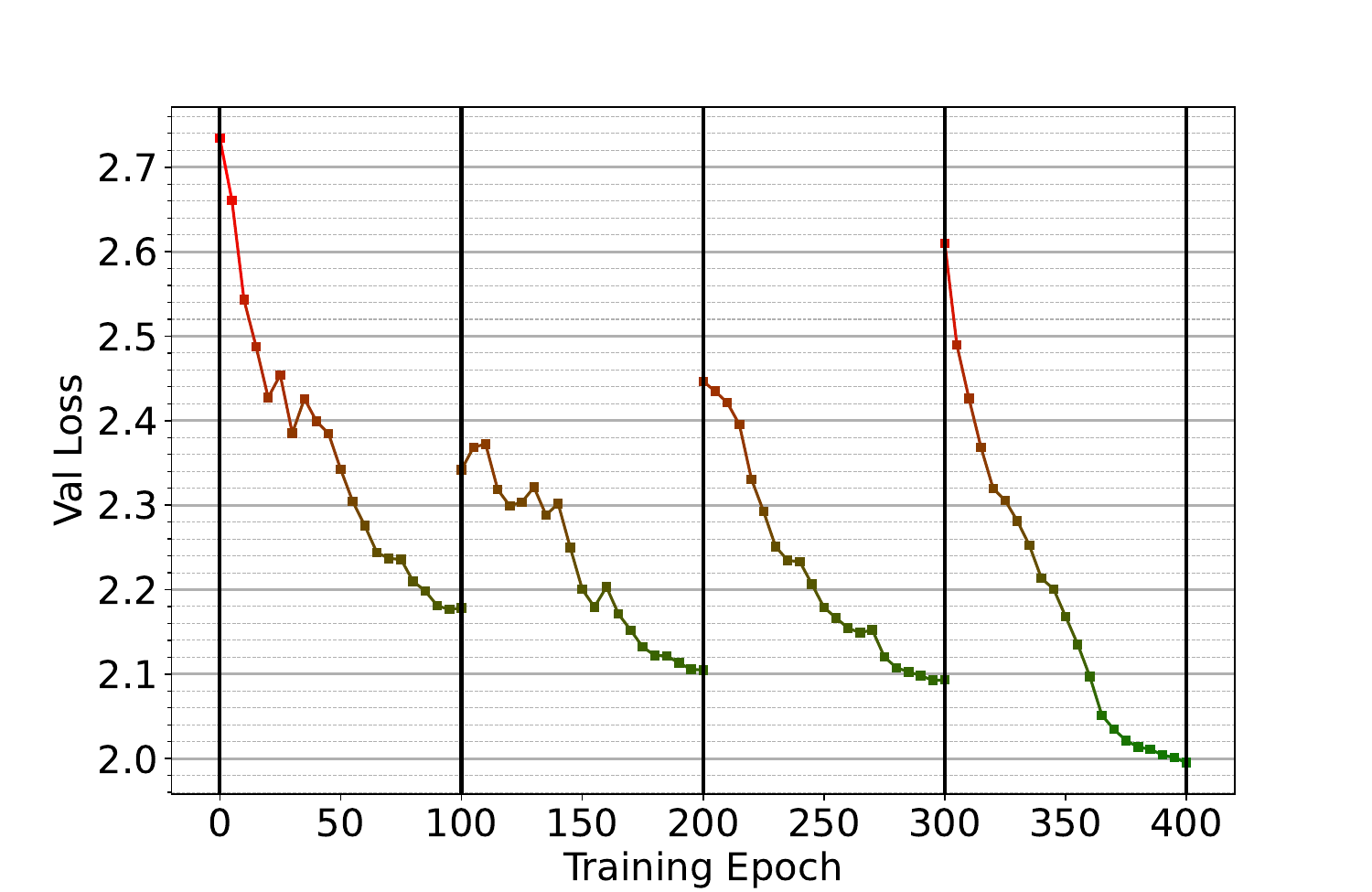}
    \vspace{-0.3cm}
    \caption{Validation loss curve for the object detection model running in the autonomous shuttle.} 
    \label{fig:ecoprt_val_loss_cascade}
    \vspace{-0.3cm}
  \end{figure}

We applied \ourmethod{} to a small-scale autonomous shuttle to continuously improve the SSD-MobileNet object detection model running in the vehicle, as shown in Figure \ref{fig:ecoprt_sys_diag}. The NVIDIA Jetson Orin device running on the shuttle uses a ZED2 stereo camera \cite{zed2} for image inputs. The system ended up taking approximately 14,000 frames at a rate of 15 frames per second, which were compressed and sent individually to the Google Cloud where we applied the periodic sampling technique with a window size of five. The vehicle's object detection model was re-trained with the ground truth generated by the Faster-RCNN model in the cloud over four retraining windows. Figure \ref{fig:ecoprt_val_loss_cascade} shows the validation loss curve for each of those retraining windows. From the final validation loss in each window, we can observe that the on-vehicle model's accuracy improves as the retraining happens. Both the SSD-MobileNet and Faster R-CNN model were initially trained on the Waymo Open Dataset. 

The cloud platform greatly accelerates the inference speed of the oracle model. For Faster R-CNN, the average inference time on the NVIDIA Jetson Orin device for one single frame captured by the shuttle reaches about 130 ms, while it takes only 28 ms for the cloud computing platform. The model training is also significantly accelerated by the cloud as well; for 10 epochs of training for the same retraining window size, it takes 422 seconds to finish on our in-lab workstation, but only 178 seconds on the cloud.

\subsection{Discussion}

\begin{itemize}

\item Vehicle data security and privacy:
Vehicles' vision data carries inherent privacy risks. 
Unauthorized access to this information could occur through network interception or compromise of cloud storage systems, potentially leading to severe breaches of individual privacy. Our usecase presented in Section~\ref{sec:usecase} ensures the privacy and integrity of the data in transit and at rest by leveraging Transport Layer Security (TLS) and storage encryption, respectively. Most cloud platforms provide these security measures. Major cloud providers also offer \emph{confidential VMs}\cite{gconfvm}, which enables the confidentiality and integrity of data \emph{in use}. This is achieved by hardware-based memory encryption; in addition to isolation among VMs, data is decrypted only when it is in use by CPU.   
This prevents other cloud tenants and even the cloud provider from accessing the data used by the confidential VMs, which can further improve the privacy of vehicle data utilized by the oracle model for ground truth generation or during the retraining process.

\item Update overhead: The lightweight nature of the device model allows us to run both the retrained and existing models simultaneously for a smooth transition between them. We executed two instances of the SSD-MobileNet model simultaneously on the NVIDIA Jetson Orin, which led to a less than 1 ms increase in the inference times for each of the model instances. Specifically, a single model instance takes an average of 4.7 ms (with a standard deviation of 0.6 ms) per frame while the concurrent execution results in the model taking 5.03 ms (with a standard deviation of 0.37 ms) per frame. Pruning, quantization, and hardware-specific optimizations help achieve these extremely low inference times.

    \item Model agnostic framework: The proposed framework is not limited to Faster R-CNN or any particular models as an oracle. Using more complex architectures like vision transformers \cite{dosovitskiy2020image} \cite{carion2020end} can provide more accurate ground truth and thus improve the retraining performance.

    \item Other ML applications: A highly accurate oracle model is crucial for leveraging the proposed framework, as it fully automates the retraining pipeline to continually improve the vehicle model. Such powerful models may not be easily available for public use in certain ML applications such as lane detection, steering angle prediction, and segmentation. Furthermore, most AV datasets are heavily focused on the object detection task, which limits our evaluation. Therefore, we limit our evaluation to object detection tasks, which are critical in AVs and for which datasets and highly accurate models are more readily available.
\end{itemize}

\section{Conclusion}

\ourmethod{} continuously improves the vision model on an autonomous vehicle by leveraging the cloud. The cloud runs a highly accurate, large oracle model that is used to guide the retraining of a small, on-vehicle perception model using the live camera stream uploaded by the vehicle. We also presented frame sampling techniques that can reduce the chance of model overfitting and also improve the prediction accuracy. An intuitive extension of this work is to expand the scale of the system by collecting image streams from multiple vehicles. This will lead to a more robust model that can generalize better to diverse novel scenarios. We leave this as future work.

\bibliographystyle{IEEEtran}
\bibliography{refs.bib}

\end{document}